\def\year{2020}\relax
\documentclass[letterpaper]{article} % DO NOT CHANGE THIS
\usepackage{aaai20}  % DO NOT CHANGE THIS
\usepackage{times}  % DO NOT CHANGE THIS
\usepackage{helvet} % DO NOT CHANGE THIS
\usepackage{courier}  % DO NOT CHANGE THIS
\usepackage[hyphens]{url}  % DO NOT CHANGE THIS
\usepackage{graphicx} % DO NOT CHANGE THIS
\usepackage{graphicx}  % DO NOT CHANGE THIS
\newtheorem{problem}{Problem}

\usepackage{amsmath}
\usepackage{amssymb}
\usepackage{bm}
\usepackage{booktabs}
\DeclareMathOperator*{\E}{\mathbb{E}}

\usepackage{flexisym}
\usepackage{multirow}
\usepackage{enumitem}
\newcommand\independent{\protect\mathpalette{\protect\independenT}{\perp}}
\def\independenT#1#2{\mathrel{\rlap{$#1#2$}\mkern2mu{#1#2}}}

\title{Minimizing Interference and Selection Bias \\in Network Experiment Design}
\author{Zahra Fatemi,
Elena Zheleva\\
Department of Computer Science, University of Illinois at Chicago\\ Chicago, IL, USA\\
\{zfatem2, ezheleva\}@uic.edu
}

\begin{document}
\maketitle
\begin{abstract}
Current approaches to A/B testing in networks focus on limiting interference, the concern that treatment effects can "spill over" from treatment nodes to control nodes and lead to biased causal effect estimation. 
Prominent methods for network experiment design rely on two-stage randomization, in which sparsely-connected clusters are identified and cluster randomization dictates the node assignment to treatment and control.   
Here, we show that cluster randomization does not ensure sufficient node randomization and it can lead to selection bias in which treatment and control nodes  represent different populations of users.
To address this problem, %By combining clustering with matching, 
we propose a principled framework for network experiment design which jointly minimizes interference and selection bias. We introduce the concepts of \emph{edge spillover probability} and \emph{cluster matching} and demonstrate their importance for designing network A/B testing. Our experiments on a number of real-world datasets show that our proposed framework leads to significantly lower error in causal effect estimation than existing solutions. 
\end{abstract}

\section{Introduction}

The gold standard for inferring causality is the use of \emph{controlled experiments}, also known as A/B tests and randomized controlled trials, in which experimenters can assign treatment (e.g. prompt users to vote) to a random subset of a population of interest and compare their outcome with the outcome of a control group, randomly selected from the same population (e.g., a group of users who were not prompted to vote). Through randomization, the experimenter can control for confounding variables that are not present in the data but can impact the treatment and outcome assignment (e.g. distance from voting polls location) and to assess whether the treatment can cause the target variable (e.g. vote) to change. Controlled experiments are widely used in the social and biological sciences~\cite{imbens-book15}, and have numerous applications in industry~\cite{varian-pnas16,kohavi-kdd13}, from understanding the impact of personalization algorithms to measuring incremental revenue due to ads. 

% 2: what are the current challenges
While it is straightforward to randomly assign treatment and control to units that are i.i.d., it is much harder to do for units that interact with each other. One of the challenges in network experiment design is dealing with interference (or spillover), the problem of treatment "spilling over" from a treated node to a control node through a shared edge, e.g., information flowing from person to person in an online social network, and diseases spreading between people interacting in the same physical space.
The presence of interference breaks the Stable Unit Treatment Value Assumption (SUTVA), the assumption that one unit's outcomes are unaffected by another unit's treatment assignment, and challenges the validity of causal inference~\cite{imbens-book15}. 
Since SUTVA is hard to guarantee in real-world scenarios, recent research on causal inference from graphs focuses on designing controlled experiments in a way that minimizes interference. 

Prominent methods for interference minimization in controlled experiments rely on graph clustering~\cite{eckles-jci17,pouget-kdd18,saveski-kdd17,ugander-kdd13}.
Graph clustering aims to find densely connected clusters of nodes, such that few edges exist across clusters~\cite{schaeffer-csr07}. The basic idea of applying it to causal inference is that little interference can occur between nodes in different clusters. Treatment and control is assigned at the cluster level, and the cluster assignment dictates the node assignment within each cluster, an experiment design known as two-stage randomization~\cite{basse-jasa18}. 

In this work, we make the key observation that there is an inherent tradeoff between interference and selection bias in cluster-based randomization based on the chosen number of clusters (as demonstrated in Figure \ref{fig:tradeoff}).
Due to the heterogeneity of real-world graphs, discovered clusters can be very different from each other, and the nodes in these clusters may not represent the same underlying population. For example, a treatment cluster may represent "predominantly Democrats from Oklahoma" while a control cluster may represent "predominantly Republicans from New York." Therefore, cluster randomization can lead to selection bias in the data with causal effects that are confounded by the difference in node features of each cluster, rather than the presence or absence of a treatment. %We demonstrate the presence of selection bias in two-stage randomization in the experimental section.
Ideally, treatment and control groups should represent the same populations, e.g., there should be clusters of "predominantly Democrats from Oklahoma" assigned to both treatment and control. A common method for dealing with selection bias in observational treatment and control data is matching, where nodes are matched based on their similarity and then assigned randomly to treatment and control~\cite{stuart-ss10}. However, there are no methods for incorporating node matching into matching graph clusters, and our work is the first to propose such a method.

Our second main contribution is in introducing the concept of "edge spillover probability" and account for it in the design. Clustering a connected graph component is guaranteed to leave edges between clusters, therefore removing interference completely is impossible. At the same time, some node pairs are more likely to interact than others and assigning such pairs to different treatment groups is more likely to lead to undesired spillover (and biased causal effect estimation) than separating pairs with low probability of interaction.

The goal of our work is to develop a framework for network experiment design in a way that minimizes both selection bias and interference.
We propose \emph{CMatch}, a two-stage framework that achieves this goal through a novel objective function for matching clusters and combining node matching with weighted graph clustering. While the idea of using graph clustering to address interference is not new~\cite{ugander-kdd13,saveski-kdd17,eckles-jci17}, incorporating node matching and edge spillover probabilities into it is novel. 

\section{Related Work}
\label{related}

Causal inference models are studied by multiple disciplines, including statistics~\cite{imbens-book15}, computer science~\cite{pearl-book09}, and the social sciences~\cite{varian-pnas16}. Here, we review relevant work on causal inference for graphs. 

\textbf{Graph mining}. Our framework relies on three preprocessing steps which can leverage existing 
graph mining algorithms for edge strength estimation, graph clustering, and node representation learning. The goal of edge (or tie or relationship) strength estimation is to assign each existing edge a numeric value which corresponds to a metric of interest for a pair of nodes, such as how close of friends two people are or how likely they are to communicate. Existing approaches rely on topological proximity~\cite{liben-nowell-ist07}, supervised models on node attributes~\cite{gilbert-chi09}, or latent variable models~\cite{li-www10}. %In our case, we would be interested to estimate how likely is treatment information to flow from one node to another. 
Graph clustering aims to find subgraph clusters with high intra-cluster and low inter-cluster edge density~\cite{yang-kis15,zhou-pvldb09}. A number of algorithms exist for weighted graph clustering~\cite{schaeffer-csr07}.
Node representation learning approaches range from graph motifs~\cite{milo-science02} to embedding representations~\cite{hamilton-tkde17} and statistical relational learning (SRL)~\cite{rossi-jair12}.

\textbf{Dealing with interference bias}.  
Recent work that addresses interference in graphs relies on separating data samples through graph clustering~\cite{backstrom-www11,eckles-jci17,gui-www15,pouget-kdd18,saveski-kdd17,ugander-kdd13}, relational d-separation~\cite{lee-aaai16,maier-aaai10,maier-uai13,marazopoulou-uai15,rattigan-aaai11}, 
or sequential randomization design~\cite{toulis-icml13}.
Since our work is closest to the approaches based on graph clustering, we use these approaches as baselines in our experiments. None of the existing approaches account for interference heterogeneity and the fact that different edges can have different spillover probabilities; this is one of our main contributions.

\textbf{Matching and selection bias}. In controlled experiments, the treatment assignment is randomized by the experimenter, whereas in estimating causal effects from observational data, the process by which the treatment is assigned is not decided by the experimenter and is often unknown. Matching is a prominent method for mimicking randomization in observational data by pairing treated units with similar untreated units. Then, the causal effect of interest is estimated based on the matched pairs, rather than the full set of units present in the data, thus reducing the selection bias in observational data~\cite{stuart-ss10}. 
There are two main approaches to matching, fully blocked and propensity score matching (PSM)~\cite{stuart-ss10}. Fully blocked matching selects pairs of units whose distance in covariate space is under a pre-determined distance threshold. 
PSM models the treatment variable based on the observed covariates and matches units which have the same likelihood of treatment. 
The few research articles that look at the problem of matching for relational domains~\cite{oktay-soma10,arbour-uai14} consider SRL data representations. None of them consider cluster matching for two-stage design which is one of our contributions. 

\section{Network experiment design}
\label{problem}

The goal of designing \emph{network experiments} is to ensure reliable causal effect estimation in controlled experiments for graphs. As a running example, imagine that we are interested to test whether showing a social media post about the benefits of voting would lead to a higher voter turnout. In this section, we present our data model, give a brief overview of the potential outcomes framework for causal inference, and present the challenges with causal effect estimation in graphs. Then, we describe the causal effects of interest and formally define the problem of network experiment design.

\subsection{Data model}

A graph $G=(V,E)$ consists of a set of $n$ nodes $V$ and a set of edges $E=\{ e_{ij}\}$ where $e_{ij}$ denotes that there is an edge between node $v_i\in V$ and node $v_j\in V$. Let $N_i$ denote the set of neighbors for node $v_i$, i.e. set of nodes that share an edge with $v_i$. 
Let $v_i.X$ denote the pre-treatment node feature variables (e.g., Twitter user features) for unit $v_i$. 
Let $v_i.Y$ denote the outcome variable of interest for each node $v_i$ (e.g., voting), and $v_i.T\in\{0,1\}$ denote whether node $v_i$ (e.g., social media user) has been treated (e.g., shown a post about the benefits of voting), $v_i.T=1$,  or not, $v_i.T=0$. $V_1$ and $V_0$ indicate the sets of units in treatment and control groups, respectively. For simplicity, we assume that both $v_i.T$ and $v_i.Y$ are binary variables. 
The edge spillover probability $e_{ij}.p$ refers to the probability of interference occurring between two nodes.

\subsection{Potential outcomes framework}
The fundamental problem of causal inference is that we can observe the outcome of a target variable for an individual $v_i$ in either the treatment or control group but not in both. Let $v_i.y(1)$ and $v_i.y(0)$ denote the \emph{potential outcomes} of $v_i.y$ if unit $v_i$ were assigned to the treatment or control group, respectively. 
The treatment effect (or causal effect) is the difference $g(i)=v_i.y(1)-v_i.y(0)$. Since we can never observe the outcome of a unit under both treatment and control simultaneously, the effect $\hat{\mu}$ of a treatment on an outcome is typically calculated through averaging outcomes over treatment and control groups via difference-in-means: $\hat{\mu}=\overline{V_1.Y}-\overline{V_0.Y}$~\cite{stuart-ss10}. 
For the treatment effect to be estimable, the following \emph{identifiability} assumptions have to hold:

\begin{itemize}[noitemsep,leftmargin=5mm]
\item \emph{Stable unit treatment value assumption} (SUTVA) refers to the assumption that the outcomes $v_i.y(1)$ and $v_i.y(0)$ are independent of the treatment assignment of other units: $\{v_i.y(1), v_i.y(0)\} \independent v_j.T, \forall v_j\neq v_i \in V$.
\item \emph{Ignorability}~\cite{imbens-book15} -- also known as \emph{conditional independence}~\cite{pearl-book09} and \emph{absence of unmeasured confoundness} %~\cite{ho-pa07} 
-- is the assumption that all variables $v_i.X$ that can influence both the treatment and outcome $v_i.Y$ are observed in the data and there are no unmeasured confounding variables that can cause changes in both the treatment and the outcome: $\{v_i.y(1), v_i.y(0)\} \independent v_i.T \mid v_i.X$. 
\item \emph{Overlap} is the assumption that each unit assigned to the treatment or control group could have been assigned to the other group. This is also known as \emph{positivity} assumption: $P(v_i.T|v_i.X)>0$ for all units and all possible $T$ and $X$.
%\consistency
\end{itemize}

\subsection{Challenges with causal effect estimation in graphs}

\begin{figure}
\centering
  \includegraphics[width=0.8\linewidth]{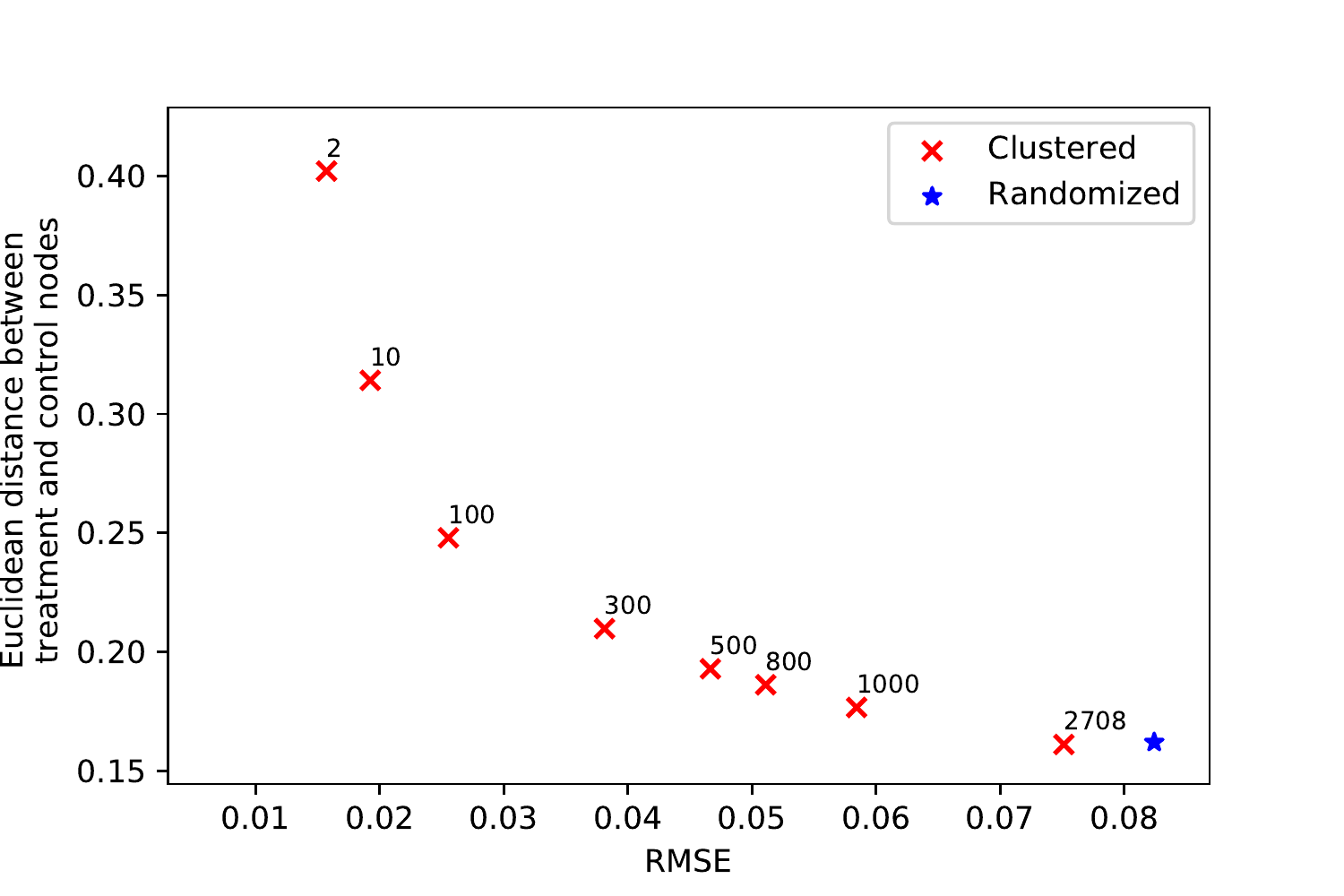}
\caption{The tradeoff between selection bias (distance) and undesired spillover (RMSE) in cluster-based randomization; each data point is annotated with the number of clusters.}
\label{fig:tradeoff}
\end{figure}

\textbf{Challenge no. 1: It is hard to separate a graph into treatment and control nodes without leaving edges across}. The presence of interference breaks the SUTVA assumption and leads to biased causal effect estimation in relational data. 
Two-stage experimental design addresses this problem by finding groups of units that are unlikely to interact with each other (stage 1) and then randomly assigning each group to treatment and control (stage 2).
Clustering has been proposed as a way to discover such groups that are strongly connected within but loosely connected across, thus finding treatment and control subgraphs that have low probability of spillover from one to the other. However, due to the density of real-world graphs, graph clustering techniques can leave as many as $65\%$ to $79\%$ of edges as inter-cluster edges (Table 2 in ~\cite{saveski-kdd17}). Leaving these edges across treatment and control nodes would lead to a large amount of spillover. % is equivalent to assuming that no information will flow through these edges which is not realistic, given the large number. 
Incorporating information about the edge probability of spillover into the clustering helps alleviate this problem and is one of the main contributions of our work.

\textbf{Challenge no. 2: There is a tradeoff between interference and selection bias in cluster-based network experiments}. While randomization of i.i.d. units  in controlled experiments can guarantee ignorability and overlap, two-stage design does not. One of the key observations in our work is that dependent on the number of clusters, there is a tradeoff between interference and selection bias in terms of the treatment and control group not representing the same underlying distribution. Figure \ref{fig:tradeoff} illustrates this tradeoff for Cora, one of the datasets in our experiments, using $reLDG$ as the clustering method. When a network is separated into very few clusters, the Euclidean distance between nodes in treatment and control clusters is larger than the Euclidean distance when a lot of clusters are produced over the same network (e.g., $0.4$ vs. $0.18$ for $2$ and $1000$ clusters). This is intuitive because as the clusters get smaller and smaller, their randomization gets closer to mimicking full node randomization (shown as a star).
At the same time, a larger number of clusters translates to a higher likelihood of edges between treatment and control nodes, which leads to higher undesired spillover and causal effect estimation error (e.g., $0.015$ vs. $0.059$ for $2$ and $1000$ clusters).

\subsection{Types of causal effects in networks}

In real-world scenarios, we are interested in estimating the \textit{Total Treatment Effect (TTE)}. 
Let $\mathbf{Z} \in\{0,1\}^N$ be the treatment assignment vector of all nodes. 
\emph{TTE} is defined as the outcome difference between two alternative universes, one in which all nodes are assigned to treatment ($\mathbf{Z_1}=\{1\}^N$) and one in which all nodes are assigned to control ($\mathbf{Z_0}=\{0\}^N$)  \cite{ugander-kdd13,saveski-kdd17}:
\begin{equation*}
TTE=\frac{1}{N} \sum_{v_i \in V} (v_i.Y(\mathbf{Z_1})-v_i.Y(\mathbf{Z_0})).
\end{equation*}
\emph{TTE} is estimated as averages over the treatment and control group, and it accounts for two types of effects, \emph{individual effects (IE)} and \emph{peer effects (PE)}: 
\begin{equation*}
\hat{TTE}=\overline{V_1.Y}-\overline{V_0.Y}=IE(V)+PE(V_1)-PE(V_0). 
\end{equation*}

Average individual effects (\emph{IE}) reflect the difference in outcome between treated and untreated subjects which can be attributed to the treatment alone. They are estimated as:
\begin{equation*}
IE(V)=\E_{v_i\in V}[v_i.Y|v_i.T=1]-\E_{v_i\in V}[v_i.Y|v_i.T=0].
\end{equation*}
Average peer effects (\emph{PE}) reflect the difference in outcome that can be attributed to influence by other subjects in the experiment. Let $N_i.\bm{\pi}$ denote the vector of treatment assignments to node $v_i$'s neighbors $N_i$. Average \emph{PE} is estimated as having neighbors with a treatment vector:
\begin{align*}
PE(V)=&\E_{v_i\in V}[v_i.Y|v_i.T=t,N_i.\bm{\pi}] \\
-&\E_{v_i\in V}[v_i.Y|v_i.T=t,N_i=\emptyset].
\end{align*}

Here, we distinguish between two types of peer effects, \textit{allowable peer effects} (\emph{APE}) and \textit{unallowable peer effects} (\emph{UPE}). Allowable peer effects are peer effects that occur within the same treatment group, and they are a natural consequence of network interactions. For example, if a social media company wants to introduce a new feature (e.g., nudging users to vote), it would introduce that feature to all users and the total effect of the feature would include both individual and peer effects. 
Unallowable peer effects are peer effects that occur across treatment groups and contribute to undesired spillover and incorrect causal effect estimation.

For each node $v_i$ in treatment group $t$, we have two types of neighbors:
1) neighbors $N_i^t$ in the same treatment class as node $v_i$ with treatment assignment set $N_i^t.\bm{\pi}$; 2) set of neighbors in a different treatment class $N_i^{\overline{t}}$ ($\overline{t}\neq t$) with treatment assignment denoted by 
$N_i^{\overline{t}}.\bm{\pi}$. 
The \emph{APE} is defined as:
\begin{align*}
APE(V)=& \E_{v_i\in V}[v_i.Y|v_i.T=t,N_i^t.\bm{\pi}]\\
-
&\E_{v_i\in V}[v_i.Y|v_i.T=t,N_i^t=\emptyset],
\end{align*}
and the \emph{UPE} is defined as:
\begin{align*}
UPE(V)=&\E_{v_i\in V}[v_i.Y|v_i.T=t,N_i^{\overline{t}}.\bm{\pi}]\\
-
&\E_{v_i\in V}[v_i.Y|v_i.T=t,N_i^{\overline{t}}=\emptyset].
\end{align*}

Ideally, we would like to measure $TTE=IE(V)+APE(V_1)-APE(V_0)$. Due to undesired spillover in a controlled experiment, what we are able to measure instead is the overall effect that comprises of both allowable and unallowable peer effects $TTE=IE(V)+APE(V_1)-APE(V_0)+UPE(V_1)-UPE(V_0)$. Therefore, when we designing an experiment for minimum interference, we are interested in setting it up in a way that makes $UPE(V_1)=0$ and $UPE(V_0)=0$.

There are two types of pairwise interference that can occur, direct interference and contagion~\cite{ogburn-ss14}. What we have described so far is causal effect estimation for direct interference which refers to a treated node $v_i$ ($v_i.X=1$) influencing the outcome of another node $v_j$. For example, a treated social media user who sees the post decides to share it with another user who ends up voting as a result. Contagion refers to the outcome of node $v_i$ influencing another node $v_j$ to have the same outcome. For example, a social media user who votes can convince another user to vote. The above definitions of peer effect can also be defined for contagion by conditioning them on neighbor outcomes, rather than neighbor treatments. We leave this exercise to the reader.

\subsection{Problem definition}
The goal of network experiment design is to minimize both unallowable peer effects and selection bias in node assignment to treatment and control. %while \fixme{allowable peer effect can exist }. 
More formally:
\begin{problem}[Network experiment design]
\label{problem-def}

Given a graph $G = (\bm{V},\bm{E})$, a set of attributes $\bm{V.X}$ associated with each node and a set of spillover probabilities $\bm{E.P}$ associated with the graph edges, we want to construct two sets of nodes, the control nodes $\bm{V}_0\in\bm{V}$ and the treatment nodes $\bm{V}_1\in\bm{V}$ such that:

\begin{itemize}
\item[a.] $\bm{V}_0 \cap \bm{V}_1 = \emptyset$%\varnothing$
\item[b.] $|\bm{V}_0|+|\bm{V}_1|$ is maximized
%\item[c.] $\theta=\sum_{\forall v_i\in \bm{V}_0, \forall v_j\in \bm{V}_1} {e_{ij}.p}$ is minimized
\item[c.] $\theta=UPE(\bm{V}_1)-UPE(\bm{V}_0)$ is minimized
\item[d.] $\bm{V}_0\bm{.X}$ and $\bm{V}_1\bm{.X}$ are identically distributed 
\end{itemize}
\end{problem}

\begin{figure*}
\centering
  \includegraphics[width=0.8\linewidth]{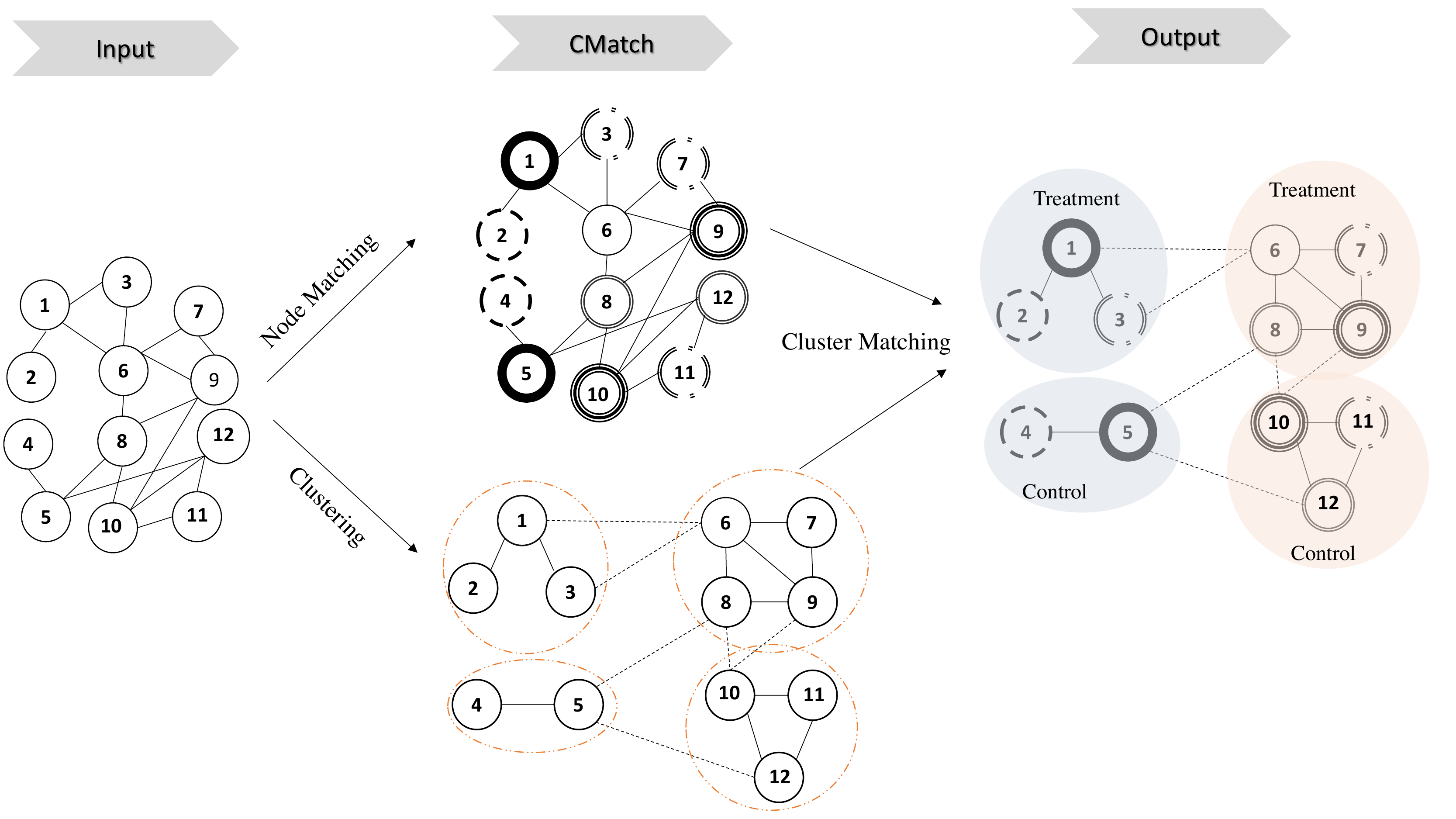}
\caption{Illustration of $CMatch$ framework for minimizing interference and selection bias in controlled experiments. \textbf{Input}: a graph of nodes and the connection between them. $\textbf{CMatch}$: node and cluster matching; the dashed circles indicates the clusters. 
Matched nodes are represented with a similar circle border.
\textbf{Output}: assigning the matched cluster pairs to treatment and control randomly; circles with the same color represent matched clusters.}
\label{fig:cmatch}
\end{figure*}

This problem definition describes the desired qualities of the experiment design at a high level.
The first component ensures that the treatment and control nodes do not overlap. The second component aims to keep as many nodes as possible from $\bm{V}$ in the final design. The third component minimizes unallowable spillover. 
The fourth component requires that there is no selection bias between the treatment and control groups. The second and third component are at odds with one another and require a trade off because the lower $\theta$, the lower the number of selected nodes for the experiment $|\bm{V}_0|+|\bm{V}_1|$. As we showed in Figure 1, there is also a tradeoff between the third and fourth component.
In the next section, we propose a solution to this problem.

\section{CMatch: a network experiment design framework}
\label{solution}
Our network experiment design framework \textit{CMatch}, illustrated in Figure \ref{fig:cmatch},  
has two main goals: 1) \textit{spillover minimization} which it achieves through weighted graph clustering, and 2) \textit{selection bias minimization} which it achieves through cluster matching. Clusters in each matched pair are assigned to different treatments, thus achieving covariate balance between treatment and control. The first goal addresses part $c$ of \textit{Problem 1} and the second goal addresses part $d$. While the first goal can be achieved with existing graph mining algorithms, solving for the second one requires developing novel approaches. To achieve the second goal, we propose an objective function, which can be solved with maximum weighted matching, and present the nuances of operationalizing each step.

\subsection{Step 1: Interference minimization through weighted graph clustering}

Existing cluster-based techniques for network experiment design assume unweighted graphs~\cite{backstrom-www11,eckles-jci17,gui-www15,saveski-kdd17,ugander-kdd13} and do not consider that different edges can have different likelihood of spillover. Incorporating information about the edge probability of spillover into the clustering helps alleviate this problem and is one of the main contributions of our work. 
In order to minimize undesired spillover, we operationalize minimizing $\theta$ as minimizing the edges, and more specifically the edge spillover probabilities, between treatment and control nodes:  $\hat{\theta}=\sum_{\forall v_i\in \bm{V}_0, \forall v_j\in \bm{V}_1} {e_{ij}.p}$.
To achieve this, $CMatch$ creates graph clusters for two-stage design by employing two functions, edge spillover probability estimation and weighted graph clustering. 

\textbf{Edge spillover probability estimation}. 
We consider edge strength, how strong the relationship between two nodes is, as a proxy for edge spillover probability. 
This reflects the notion that the probability of a person influencing a close friend to do something is higher than the probability of influencing an acquaintance. 
We can use common graph mining techniques to calculate edge strength, including ones based on topological proximity~\cite{liben-nowell-ist07}, supervised classification~\cite{gilbert-chi09}, or latent variable models~\cite{li-www10}. 

\textbf{Weighted graph clustering}.
In order to incorporate edge strength into clustering, we can use any existing weighted graph clustering algorithm~\cite{enright-nar02,schaeffer-csr07,yang-kis15}. 
In our experiments,
we use a prominent non-parametric algorithm, the \textit{Markov Clustering Algorithm (MCL)}~\cite{enright-nar02} which applies the idea of random walk for clustering graphs and produces non-overlapping clusters. We also compare this algorithm with \textit{reLDG} which was the basis of previous work~\cite{saveski-kdd17}. One of the advantages of $MCL$ is that it automatically finds the optimal number of clusters, rather than requiring it as input. 
The main idea behind $MCL$ is that nodes in the same cluster are connected with higher-weighted shortest paths than nodes in different clusters.

\subsection{Step 2: Selection bias minimization through cluster matching}
Randomizing treatment assignment over clusters in a two-stage design does not guarantee that nodes within those clusters would represent random samples of the population. We propose to address this selection bias problem by \textit{cluster matching} and balancing covariates across treatment and control clusters. While methods for matching nodes exist~\cite{stuart-ss10,oktay-soma10,arbour-uai14}, this work is the first to propose methods for matching clusters. 

\textbf{Objective function}.
The goal of cluster matching is to find pairs of clusters with similar node covariate distributions and assign them to different treatment groups. We propose to capture this through a maximum weighted matching objective over a cluster graph in which each discovered cluster from step $1$ is a node and edges between clusters represent their similarity. 
Suppose that graph $G$ is partitioned into $C=\{c_1, c_2, ..., c_g\}$ clusters.
We define $A\in\{0,1\}$, such that $a_{ij}=1$ if two clusters $c_i$ and $c_j$ are matched, else $a_{ij}=0$. $w_{i,j}\in \mathbb{R}$ represents the similarity between two clusters $c_i$ and $c_j$.
Then the objective function of $CMatch$ is as follows:
\begin{equation}
\label{objective}
\begin{aligned}
%& \underset{}{\text{maximize}} 
& \underset{A}{\arg\max}
%&& \sum_{i=1}^{g} \sum_{j=i+1}^{g} (a_{ij} \times \sum_{k=1}^{|c_i|}\sum_{l=1}^{|c_j|} r^{ij}_{kl}) \\
&& \sum_{i=1}^{g} \sum_{j=i+1}^{g} (a_{ij} \cdot w_{ij}) \\
& \text{subject to}
&& \forall c_i \in C, \sum_{j=1}^{|c_i|} a_{ij} \leq 1 , a_{ij} \in \{0,1\}.
\end{aligned}
\end{equation}

This objective function maps to a maximum weighted matching problem for which there is a linear-time approximation algorithm~\cite{duan-acm14} and a polynomial-time exact algorithm with $O(N^{2.376})$~\cite{mucha-ieee04,harvey-siam09}. 

\textbf{Solution}. In order to operationalize the solution to this objective, the main question that needs to be addressed is: what does it mean for two clusters to be similar? We propose to capture this cluster similarity through matched nodes. The more nodes can be matched based on their covariates across two clusters, the more similar two clusters are. Thus, the operationalization comes down to the following three questions which we address next: 
\begin{enumerate}
\item What constitutes a \textit{\textbf{node match}}?
\item How are node matches taken into consideration in computing the pairwise \textit{\textbf{cluster weights}} (cluster similarity)?
\item Given a cluster weight, what constitutes a potential cluster match, and thus an edge in the \textit{\textbf{cluster graph}}?
\end{enumerate}
Once these three questions are addressed, the cluster graph can be built and an existing maximum weighted matching algorithm can be applied on it to find the final cluster matches.

\textit{\textbf{Node Matching}}. The goal of node matching is to reduce the imbalance between treatment and control groups due to their different feature distributions. 
Given a node representation, fully blocked matching would look for the most similar nodes based on that representation~\cite{stuart-ss10}. It is important to note that propensity score matching does not apply here because it models the probability of treatment in observational data and treatment is unknown at the time of designing a controlled experiment.  
In its simplest form, a node can be represented as a vector of attributes, including node-specific attributes, such as demographic characteristics, and structural attributes, such as node degree. For any two nodes, it is possible to apply an appropriate similarity measure $sim(v_i,v_j)$, in order to match two nodes, including cosine similarity, Jaccard similarity or Euclidean distance.  

We consider two different options to match a pair of nodes in different clusters (and ignore matches within the same cluster):
\begin{itemize}
\item \textbf{Threshold-based node matching (TNM)}: Node $v_k$ in cluster $c_i$ is matched with node $v_l$ from a different cluster $c_j$ if the pairwise similarity of nodes $sim(v_k,v_l)>\alpha$. The threshold $\alpha$ can vary from $0$, which liberally matches all pairs of nodes, to the maximum possible similarity which matches nodes only if they are exactly the same. In our experiments, we set $\alpha$ based on the covariate distribution of each dataset and consider different quartiles of pairwise similarity as thresholds. This allows for each node to have multiple possible matches across clusters.
\item \textbf{Best node matching (BNM)}: Node $v_k$ in cluster $c_i$ is matched with only one node $v_l$ which is most similar to $v_k$ in the whole graph; $v_l$ should be in a different cluster. This is a very conservative matching approach in which each node is uniquely matched but allows the matching to be asymmetric.
\end{itemize}  

\textit{\textbf{Cluster Weights}}. 
After the selection of a node matching mechanism, we are ready to define the pairwise similarity of clusters which is the basis of cluster matching.  We consider three simple approaches and three more expensive approaches which require maximum weighted matching between nodes:

\begin{itemize}
\item \textbf{Euclidean distance (E)}: This approach is the simplest of all because it does not consider node matches and it simply calculates the Euclidean distance between the node attribute vector means of two clusters. 
\item \textbf{Matched node count (C)}: The first approach counts the number of matched nodes in each pair of clusters $c_i$ and $c_j$ and consider the count as the clusters' pairwise similarity: $w_{ij}=\sum_{k=1}^{|c_i|}\sum_{l=1}^{|c_j|} r^{ij}_{kl}$. A node in cluster $c_i$ can have multiple matched nodes in $c_j$. 
\item \textbf{Matched node average similarity (S)}: Instead of the count, this approach considers the average similarity between matched nodes across two clusters $c_i$ and $c_j$: \\$w_{ij}=\frac{\sum_{k=1}^{|c_i|}\sum_{l=1}^{|c_j|} r^{ij}_{kl} \cdot sim(v_k,v_l)}{\sum_{k=1}^{|c_i|}\sum_{l=1}^{|c_j|} r^{ij}_{kl}}$.
\end{itemize}

These first two approaches allow a single node to be matched with multiple nodes in another cluster and each of those matches to count towards the cluster pair weight. In order to distinguish this from a more desirable case in which multiple nodes in one cluster are matched to multiple nodes in another cluster, we propose approaches that allow each node to be considered only once in the matches that count towards the weight. For each pair of clusters, we build a node graph in which an edge is formed between nodes
%\fixme{we build a bipartite node graph in which an edge is formed between matched nodes}
$v_i$ and $v_j$ in the two clusters and the weight of this edge is $sim(v_i,v_j)$. Maximum weighted matching will find the best possible node matches in the two clusters. We consider three different variants for calculating the cluster pair weight based on the maximum weighted matching of nodes:

\begin{itemize}
\item \textbf{Maximum matched node count (MC)}: This method calculates the cluster weight the same way as \textbf{C} except that the matches (whether $r^{ij}_{kl}$ is $0$ or $1$) are based on the maximum weighted matching result. 

\item \textbf{Maximum matched node average similarity (MS)}: This method calculates the cluster weight the same way as \textbf{S} except that the node matches are based on the maximum weighted matching result. 

\item \textbf{Maximum matched node similarity sum (MSS)}: This method calculates the cluster weight similarly to \textbf{MS} except that it does not average the node similarity: $w_{ij}=\sum_{k=1}^{|c_i|}\sum_{l=1}^{|c_j|} r^{ij}_{kl} \cdot sim(v_k,v_l)$.

\end{itemize}    

\textit{\textbf{Cluster Graph}}. Once the cluster similarities of have been determined, we need to decide what similarity constitutes a potential cluster match. Such potential matches are added as edges in the cluster graph which is considered for maximum weighted matching. We consider three different options:

\begin{itemize}
\item \textbf{Threshold-based cluster matching (TCM)}: Cluster $c_i$ is considered as a potential match of cluster $c_j$ if their weight $w_{i,j}>\beta$. The threshold $\beta$ can vary from $0$, which allows all pairs of clusters to be potential matches, to the maximum possible similarity which allows matching between clusters only if they are exactly the same. In our experiments, we set $\beta$ based on the distribution of pairwise similarities and their quartiles as thresholds. 

\item \textbf{Greedy cluster matching (GCM)}: For each cluster $c_i$, a sorted list of the similarities between $c_i$ and all other clusters is defined. Cluster $c_i$ is considered a potential match only to the cluster with the highest similarity value in the list.
\end{itemize}

The last step in $CMatch$ runs maximum weighted matching on the cluster graph. For every matched cluster pair, it assigns one cluster to treatment and the other one to control at random. This completes the network experiment design.

%\begin{figure*}
%\centering
%  \includegraphics[width=1\linewidth]{fig/RMSE_contagion.JPG}
%\caption{RMSE of total effect in the presence of contagion considering different spillover probabilities}
%\label{fig:CausalPlot}
%\vspace{-5pt}
%\end{figure*}

\section{Experiments}
\label{experiments}
In this section, we evaluate the advantages of $CMatch$ in minimizing interference and selection bias compared to state-of-the-art methods. % We measure 1) the error of total effect estimation and the number of edges between treatment and control nodes to show the strength of interference bias and, 2) the distance between treatment and control nodes' attributes to represent the size of selection bias in different estimators.
% We only considered attributed graph datasets because we need node attributes for calculating node similarity.

\subsection{Semi-synthetic data}
We consider four real-world attributed network datasets. %showing the advantages of $CMatch$ over state-of-the-art methods for causal effect estimation.
Table \ref{table:datasets} shows the basic dataset characteristics. The \emph{50 Women} dataset ~\cite{michell-ssm97} incorporates the smoking, alcohol, sport and drug habits of $50$ students. $Hamsterster$ ~\cite{zheleva-snakdd08} describes the online social network of hamsters and their attributes. $Cora$ and $Citeseer$ ~\cite{sen-aimag08} are two citation networks with binary bag-of-words attributes for each article. %We use the attributes of datasets to measure the similarity of their nodes. 

% In order to explore the impact of different potential interference mechanisms on estimated causal effects, consider direct interference where a treated neighbor can activate a control node and contagion where an activated treatment node can activate a controlled node and vice versa with the (undesired) edge spillover probability $e.p$.

We assume that the underlying probability of activating a node (changing the outcome) due to treatment and allowable peer effects in the treatment group is $0.4$ and the underlying probability of activating a control node due to treatment and allowable peer effects is $0.2$ which makes the true causal effect $TTE=0.2$.
%In our experiments, we active $40\%$ of treatment and $20\%$ of control nodes randomly. 
Based on these probabilities, we randomly assign each node as activated or not.
For each inactivated nodes, we simulate two types of interference considering both fixed values ($0.1$ and $0.5$) and values based on the edge weights for $e.p$:
\begin{enumerate}
    \item Direct interference: each treated neighbor of a control node activates the node with unallowable spillover probability of $e.p$. 
    \item Contagion: inactive treated and untreated nodes get activated with the unallowable spillover probability of $e.p$ if they are connected to at least one activated node in a different treatment class.
\end{enumerate}

%\subsection{Algorithms for preprocessing the data}
%\fixme{} 

\begin{table}[ht]

\centering
\caption{Number of nodes, edges and attributes in the  datasets}
%\small\addtolength{\tabcolsep}{-4pt}
\begin{tabular}{lrrr}  
\toprule
 Dataset & Nodes & Edges & Attributes\\
\midrule
 %Catster &73204  & 221293 \\
 Citeseer &3,312&4,675&3709\\
 Cora &2,708 & 5,278 &1440\\
 Hamsterster & 2,059  & 10,943 &6\\
 50 Women & 50 &74& 4 \\
 
 %Hateful Users~\cite{ribeiro-icwsm18} &100,386  & 2,194,979 &1024\\
\bottomrule
\end{tabular}

\label{table:datasets}
\end{table}

\subsection{Main algorithms and baselines}
% We used two baselines from state-of-the-art approaches to relational causal inference ($\bm{CR}$, $\bm{CBR_{reLDG}}$)~\cite{saveski-kdd17}, one hybrid baseline which incorporates weighted clustering into $\bm{CBR}$ ($\bm{CBR_{MCL}}$), one baseline which considers matching but ignores interference ($\bm{Match}$), and two versions of our algorithm, one that considers an unweighted graph ($\bm{CMatch_{reLDG}}$) and a fully-fledged one that incorporates edge weights ($\bm{CMatch_{MCL}}$). 
All our baseline and main algorithm variants take an attributed graph as an input and produce a set of clusters, each assigned to treatment, control, or none.
For graph clustering, we considered two main algorithms, \emph{Restreaming Linear Deterministic Greedy (reLDG)}~\cite{nishimura-kdd13} and \emph{Markov Clustering Algorithm (MCL)}~\cite{enright-nar02}. 
%\fixme{and \emph{Bayesian Attributed Graph Clustering (BAGC)}~\cite{xu-sigmod12}}.
$reLDG$ takes as input an unweighted graph and desired number of clusters and produces a graph clustering. $reLDG$ was reported to perform very well in state-of-the-art methods for network experiment design~\cite{saveski-kdd17}. $MCL$ is a non-parametric algorithm which takes as input a weighted graph and produces a graph clustering. The edge weights which correspond to the probabilities of spillover are estimated based on node pair similarity
using one minus the normalized L2 norm: $1-L_2(v_i.x,v_j.x)$. %Here, we list the main algorithms we tested, and after that provide more details on the baselines from ~\cite{saveski-kdd17}:

%$BAGC$ is a model based approach which uses structural and attribute information of an unweighted gra to partition it to the desired number of clusters.}

The main algorithms and baselines are:
    \begin{itemize}
        \item $\bm{Randomized}$: This algorithm assigns nodes to treatment and control randomly, ignoring the network.
        \item $\bm{CR}$~\cite{saveski-kdd17}: The \emph{Completely Randomized (CR)} algorithm was used as a baseline in ~\cite{saveski-kdd17}. The algorithm clusters the unweighted graph using \emph{reLDG} algorithm, assigns similar clusters to the same strata and assigns nodes in strata to treatment and control in a randomized fashion
        %creates cluster strata based on cluster attributes and assigns nodes in strata to treatment and control in a randomized fashion. %Clustering the graph, strata assignment in cluster-level and treatment assignment in the node level are the main steps.
        \item $\bm{CBR_{reLDG}}$~\cite{saveski-kdd17}: \emph{Cluster-based Randomized assignment (CBR)} is the main algorithm proposed by ~\cite{saveski-kdd17}. The algorithm clusters the unweighted graph using \emph{reLDG}, assigns similar clusters to the same strata and randomly picks clusters within the same strata as treatment or control. % \cite{saveski-kdd17} experimented with a number of graph clustering algorithms, of which the best-performing algorithm was \emph{reLDG} which is why we included the \emph{reLDG} variant of \emph{CBR}.
        \item $\bm{CBR_{MCL}}$: A variant of $\bm{CBR}$ that we introduce for the sake of fairness which uses $MCL$ for weighted-graph clustering. %The main difference between $CBR_{MCL}$ and $CBR_{reLDG}$ is that $CBR_{MCL}$ uses $MCL$ for graph clustering instead of $reLDG$.
        \item $\bm{Match}$: This algorithm matches nodes using maximum weighted matching algorithm and then randomly assigns nodes in each matched pair to treatment and control at random without considering clustering. 
        %There is no clustering in this method. %The point is that $Match$ first matches nodes and then assigns one node in each matched node pair to treatment and the other one to control at random. 
        \item $\bm{CMatch_{reLDG}}$: This method uses our $CMatch$ framework but works on an unweighted graph. It uses $reLDG$ for graph clustering. %$CMatch_{reLDG}$ shows the importance of jointly optimizing node matching and clustering for causal effect estimation in graphs but without incorporating information flow likelihood.
        \item $\bm{CMatch_{MCL}}$: This is our proposed technique which uses $MCL$ for weighted graph clustering. %preprocesses the weighted graph by applying graph clustering with $MCL$ and matching the nodes. Then, it runs the $CMatch$ optimization algorithm to match clusters and assign one cluster in each matched cluster pair to treatment and the other one to control at random. % (based on degree similarity or Euclidean distance, count the number of matched nodes in each two cluster and assigning each of matched clusters to treatment or control randomly.
        %\fixme{
        %\item $\bm{CMatch_{BAGC}}$: This method uses $CMatch$ technique and $BAGC$ as the clustering algorithm.
       % }
    \end{itemize}

$CMatch$ uses the $maximum\_weight\_matching$ function from the $NetworkX$ Python library. 

%For the $CMatch$ implementations, we used the the maximum weight matching function from the $NetworkX$ Python library to calculate the best \fixme{node and} cluster matches in each dataset. 
% \fixme{We also calculate the number of edges and sum of edge weights between treatment and control nodes using attribute matching with Euclidean distance as the matching technique. For each node, we calculate the distance between the node and all other nodes in graph and match it with the most similar node if it is not matched before. Each node is matched with at most one node. In cluster-level matching, we count the number of matched nodes between clusters and match each two clusters with maximum number of matched nodes.}

\begin{figure}
\centering
  \includegraphics[width=1\linewidth]{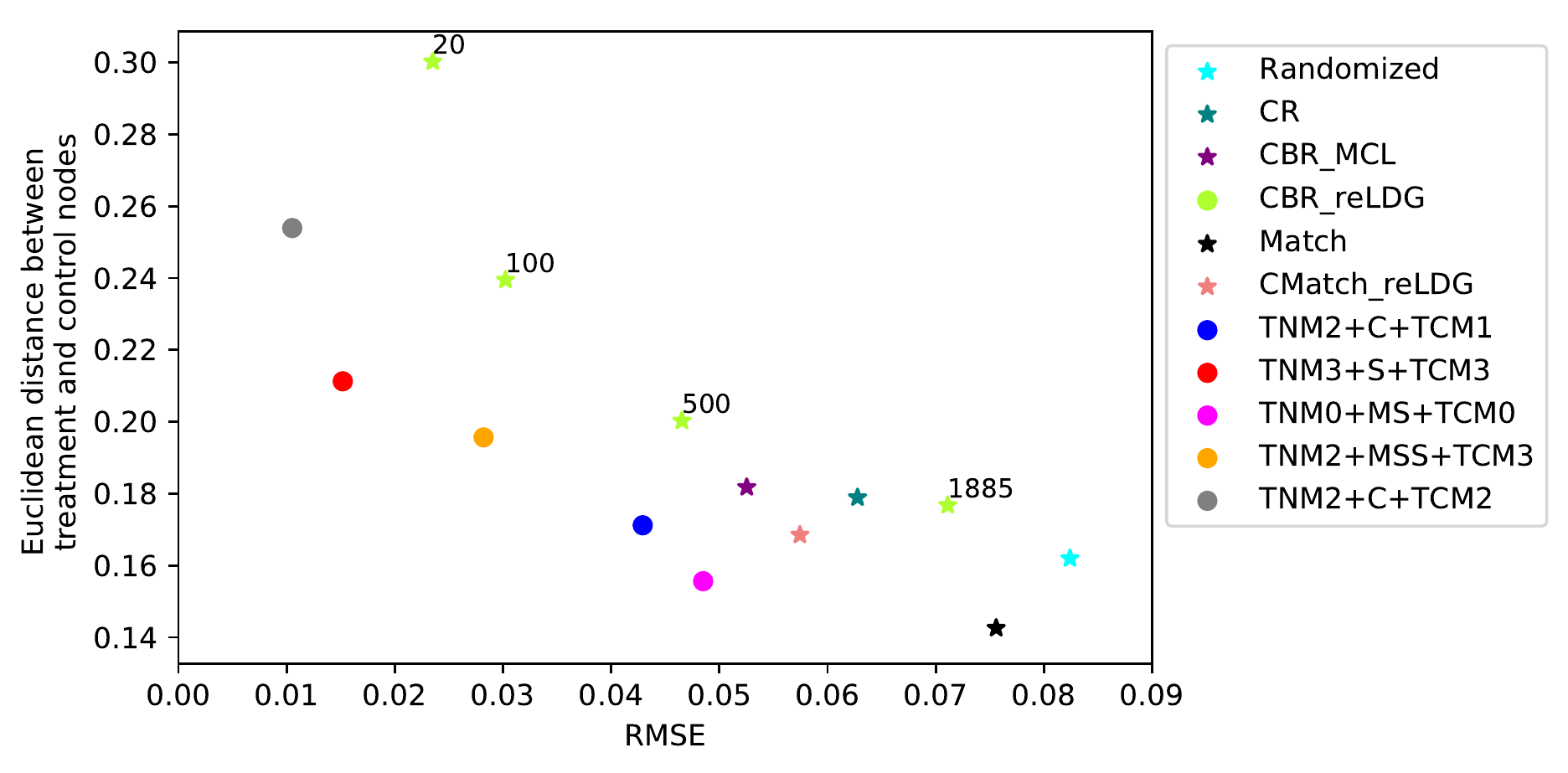}
\caption{The tradeoff between selection bias (distance) and undesirable spillover (RMSE) in $CMatch_{MCL}$ variants (labeled with methods applied in) and baselines in Cora dataset for \emph{e.p = edge-weight}; $CBR_{reLDG}$ is annotated with the number of clusters}
\label{fig:cmatch_tradeoff}
\end{figure}

\begin{figure*}
\centering
  \includegraphics[width=1\linewidth]{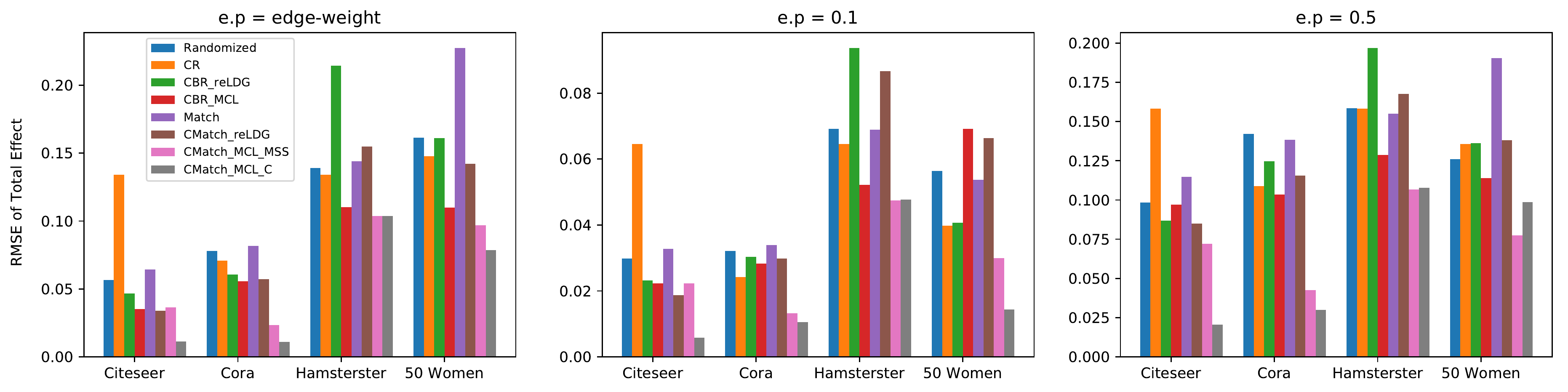}
\caption{RMSE of total effect in the presence of contagion considering different unallowable spillover probabilities in all datasets; $CMatch_{{MCL}_{C}}$ achieves the lowest error in all datasets}
\label{fig:CausalPlot}
\end{figure*}

\subsection{Experimental setup}
We run a number of experiments varying the underlying spillover assumptions, clustering algorithms, number of clusters, and node matching algorithms. Our experimental setup measures the desired properties for network experiment design, as described in Problem~\ref{problem-def} and follows the experimental setups in existing work~\cite{arbour-uai14,eckles-jci17,maier-uai13,stuart-ss10,saveski-kdd17}. % Due to space limits, we describe only the most insightful ones.
 
To measure the strength of interference bias in different estimators, we report on two metrics:  
 \begin{enumerate}
     \item 
% To show the accuracy of different methods in estimating causal effect,
 \textit{Root Mean Squared Error} (RMSE) of the total treatment effect calculated as:
\begin{equation*}
   RMSE= \sqrt{\frac{1}{S}\sum_{s=1}^{S} ((\hat{\tau}_s-\tau_s)^2)}
\end{equation*}
where $S$ is the number of runs and $\tau_s$ and $\hat{\tau}_s$ are the true and estimated causal effect in run $s$, respectively. We set $S=10$ in all experiments. Error can be attributed to undesired spillover only.
\item The number of edges and sum of edge weights between treatment and control nodes assigned by each algorithm.
%as proxies for measuring interference bias.
\end{enumerate}
To show the selection bias, we want to assess how different treatment vs. control nodes are. We compute the Euclidean distance between the attribute vector mean of treated and that of untreated nodes. %' attributes for different number of clusters in three different estimators and to show the statistical significance of the estimates, 
We show the average and standard deviation over 10 runs. 

We run all $115$ possible combinations of CMatch options for node matching, cluster weights and cluster graph for each dataset. %, we show the tradeoff between potential spillover and selection bias in some variants. 
We consider four different values for the threshold $\alpha$ in \textbf{TNM}: 0 (\textbf{TNM0}), first (\textbf{TNM1}), second (\textbf{TNM2}) and third (\textbf{TNM3}) quantile of pairwise nodes' similarity distribution where $sim(v_i,v_j)$= (1- the normalized $L_2$ norm). For \textbf{TCM}, we consider four different $\beta$ values: 0 (\textbf{TCM0}), first (\textbf{TCM1}), second (\textbf{TCM2}) and third (\textbf{TCM3}) quantile of the pairwise clusters' similarity distribution for each dataset.
We use \textbf{TNM2 + C + TCM2} in all the experiments of $CMatch_{reLDG}$. 

Unless otherwise specified, the number of clusters is the same for all CBR and CMatch versions based on the optimal determination by MCL as optimal for each respective dataset. The number of clusters determined by $MCL$ is $2,497$ for $Citeseer$, $1,885$ for $Cora$, $1,056$ for $Hamsterster$ and $20$ in \emph{50 Women} dataset.

\begin{table*}[h]
 %\small\addtolength{\tabcolsep}{-2pt}
 %\columnwidth
 \caption{The tradeoff between selection bias (distance) and undesirable spillover (RMSE) in $CMatch$ variants in Cora dataset. $CMatch_{MCL}$ variants used in Fig. \ref{fig:cmatch_tradeoff} are in bold.}
\centering
    \begin{tabular}{|c|c|c|c|c|c|c|c|c|c|c|c|}
    \hline
    &&\multicolumn{2}{|c|}{TCM0}&\multicolumn{2}{c}{TCM1}&\multicolumn{2}{|c|}{TCM2}&\multicolumn{2}{|c|}{TCM3}&\multicolumn{2}{|c|}{GCM}\\ \hline
    &&RMSE&ED&RMSE&ED&RMSE&ED&RMSE&ED&RMSE&ED\\
    \hline
    \multirow{5}{*}{C}&TNM0&0.052&0.184&0.007&0.267&0.017&0.263&0.014&0.26&0.048&0.789\\
    &TNM1&0.055&0.176&0.051&0.177&0.008&0.258&0.012&0.26&0.031&0.6\\
    &TNM2&0.054&0.171&\textbf{0.042}&\textbf{0.171}&\textbf{0.01}&\textbf{0.253}&0.017&0.251&0.036&0.591\\
    &TNM3&0.043&0.175&0.043&0.175&0.173&0.046&0.018&0.231&0.034&0.592\\
    &BNM&0.012&0.262&0.037&0.481&0.049&0.485&0.059&0.479&0.025&0.274\\
    \hline
    \multirow{5}{*}{S}&TNM0&0.056&0.16&0.058&0.159&0.048&0.16&0.056&0.162&0.035&0.34\\
    &TNM1&0.055&0.16&0.053&0.162&0.057&0.165&0.054&0.166&0.026&0.31\\
    &TNM2&0.056&0.162&0.054&0.168&0.048&0.165&0.033&0.183&0.039&0.292\\
    &TNM3&0.057&0.169&0.041&0.174&0.024&0.198&\textbf{0.015}&\textbf{0.211}&0.021&0.275\\
    &BNM&0.014&0.253&0.017&0.264&0.02&0.27&0.027&0.303&0.014&0.277\\
    \hline
    \multirow{5}{*}{MC}&TNM0&0.049&0.177&0.015&0.261&0.01& 0.262&0.008&0.263& 0.042&0.189\\
    &TNM1&0.055&0.173&0.052&0.174&0.01&0.257&0.012&0.253 &0.040&0.191\\
    &TNM2&0.047&0.171&0.051&0.177&0.013&0.261&0.007&0.263&0.024&0.211\\
    &TNM3&0.047&0.173&0.049&0.178&0.051&0.176&0.011&0.249&0.012&0.244\\
    &BNM&N/A&N/A&N/A&N/A&N/A&N/A&N/A&N/A&N/A&N/A\\
    \hline
    \multirow{5}{*}{MS}&TNM0&\textbf{0.048}&\textbf{0.155}&0.051&0.156&0.052&0.156&0.058&0.157&0.018&0.271\\
    &TNM1&0.051&0.156&0.057&0.157&0.048&0.156&0.052&0.16&0.022&0.264\\
    &TNM2&0.059&0.156&0.057&0.157&0.054&0.158& 0.056&0.157& 0.021&0.258\\
    &TNM3&0.053&0.157&0.05&0.159&0.056&0.155& 0.051&0.156&0.028&0.27\\
    &BNM&N/A&N/A&N/A&N/A&N/A&N/A&N/A&N/A&N/A&N/A\\
    \hline
    \multirow{5}{*}{MSS}&TNM0&0.059&0.162&0.048&0.162&0.061&0.159& 0.036&0.184& 0.026&0.271\\
    &TNM1&0.056&0.16&0.054&0.161&0.047&0.161& 0.03&0.194& 0.029&0.275\\
    &TNM2&0.052&0.161&0.057&0.161&0.045&0.172 &\textbf{0.028}& \textbf{0.195}& 0.021&0.281\\
    &TNM3&0.049&0.168&0.035&0.186&0.023&0.199&0.022&0.212&0.033&0.278\\
    &BNM&N/A&N/A&N/A&N/A&N/A&N/A&N/A&N/A&N/A&N/A\\
    \hline
    E&N/A&0.051&0.178&0.05&0.18&0.031&0.203&0.012&0.242&0.042&0.718\\
    \hline
    \end{tabular}
    \label{table:cmatch_variants}
 \end{table*}

\subsection{Results}
% Here, we provide quantitative and qualitative analysis for different estimators from two aspects of interference and selection bias.
%we describe the main findings of comparing $CMatch$ with the state-of-art estimators from two aspects of interference and selection bias.
%provide quantitative and qual- itative analysis for different LSE methods
\textbf{Tradeoff between interference and selection bias in \emph{CMatch} variants and baselines:} 
%We explore the difference between causal effect estimation of \emph{CMatch} variants in different datasets, 
Given the large number of $CMatch$ option combinations ($115$), we first find which ones of these combinations have a good tradeoff between RMSE and Euclidean distance (between treatment and control) with \emph{e.p = edge-weight}.
Based on these experiments, we notice that 1) methods with stricter cluster thresholds (\textbf{TCM2} and \textbf{TCM3}) tend to have lower error,
2) stricter node match thresholds (\textbf{TNM2} and \textbf{TNM3}) have lower error than others for \textbf{S} and \textbf{MSS} and
3) \textbf{MS} has high error across thresholds. Due to space constraints, we are showing detailed results for Cora only.

Fig. \ref{fig:cmatch_tradeoff} shows the results %between selection and interference biases in some variants of $CMatch_{MCL}$, 
for the $CMatch$ variants with the best tradeoffs and their better performance when compared to the baselines for Cora. Full $CMatch$ results can be found in Table \ref{table:cmatch_variants}. %, labeled by $CMatch$ options applied in them, and baseline methods in Cora for $e.p$ = edge-weight. 
The figure clearly shows that the selection bias decreases at the expense of interference bias. % increase in such a way that in $CMatch$ variants by decreasing 
For example, while the Euclidean distance for \textbf{TNM0+MS+TCM0} is low ($0.155$) when compared to \textbf{TNM2+C+TCM2} ($0.253$), its RMSE is higher, $0.048$ vs. $0.01$.
%In Cora, the optimal number of clusters found by $MCL$ is $1885$ clusters. 
% All of these best variants are better
The comparison between $CBR_{reLDG}$ with different possible number of clusters %(annotated points) 
is consistent with the tradeoff shown in Fig.\ref{fig:tradeoff}. %: more clusters results in lower Euclidean distance and higher causal effect estimations error. 
$CBR_{reLDG}$ with the highest error (annotated with $1885$) and $CMatch_{MCL}$ have the same number of clusters.
It is intuitive that the $Match$ method has the least selection bias, because all nodes have their best matches. However, similar to the $Randomized$ method, it suffers from high interference bias (RMSE) because of the high density of edges between treatment and control nodes. 

\begin{figure}[h]
\centering
  \includegraphics[width=1\linewidth]{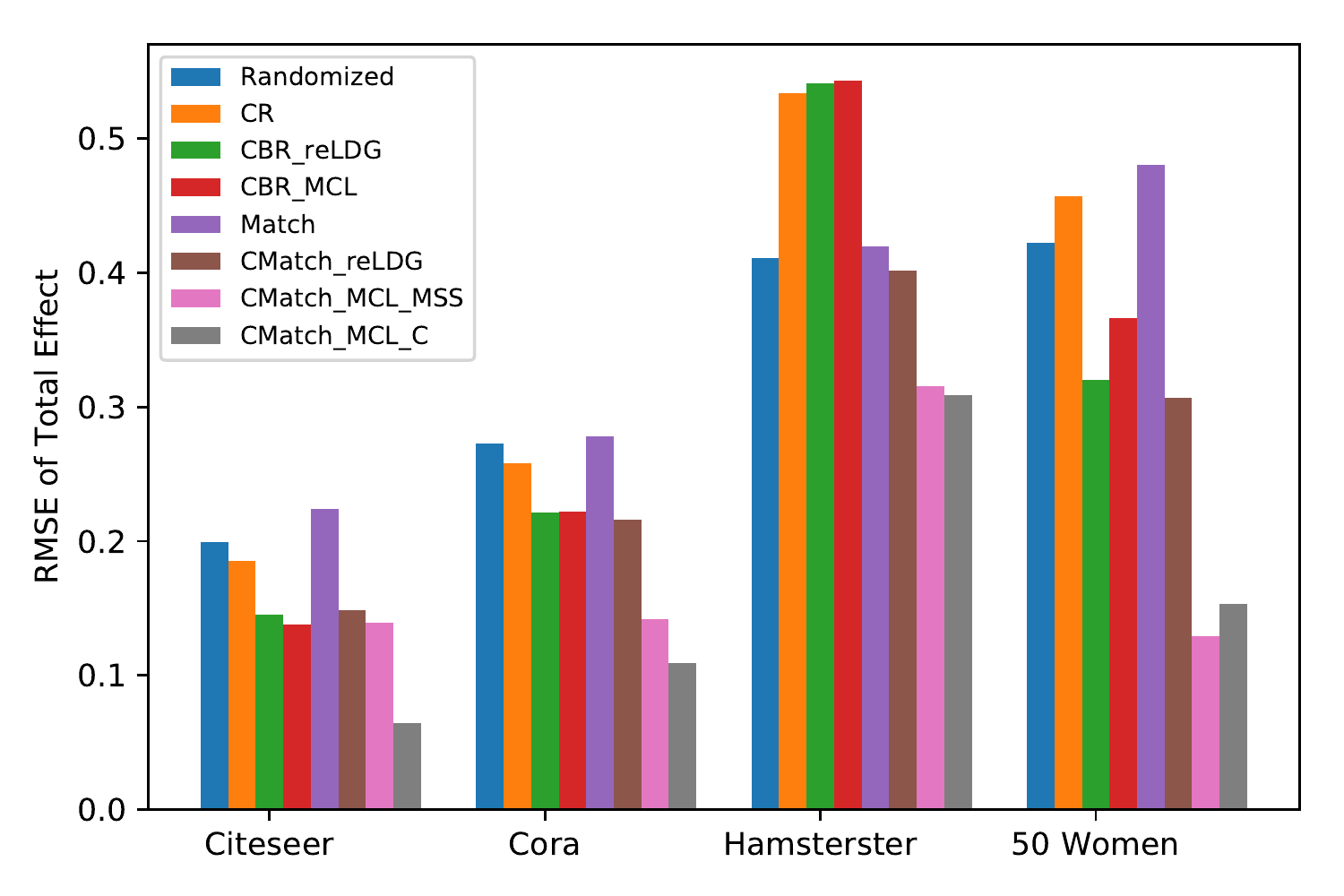}
\caption{RMSE of total effect in the presence of direct interference (\emph{e.p = edge-weight}). 
$CMatch_{{MCL}_C}$ and $CMatch_{{MCL}_MSS}$ obtain the lowest RMSE for all datasets.
}
\label{fig:directinterference}
\end{figure}

\begin{table*}
\caption{Percentage of edges (and edge weights) between treatment and control nodes.  %using degree matching technique. 
The lower the number, the lower probability of undesired spillover.}
\centering
\small\addtolength{\tabcolsep}{-2pt}
\begin{tabular}{lrrrrrrr}  
%\begin{adjustbox}{width=\textwidth}
\toprule
Dataset & $Randomized$ &$CR$& $CBR_{reLDG}$& $CBR_{MCL}$ & $Match$ &$CMatch_{reLDG}$ & $CMatch_{{MCL}_{C}}$  \\
\midrule
 Citeseer & 49.9\%(50\%) &35.9\% (36.3\%) & 39.8\% (38.4\%)&38.9\%(38.4\%)&53.9\% (56.6\%) &35.8\% (34.4\%)& $\bm{7.5\%} (7.2\%)$ \\
 Cora &49.7\%(49.7\%)&37.6\%(37.6\%) & 43.4\%(42.8\%) & 38.9\% (33.6\%)& 51.8\%(53.3\%)&  38.7\% (38.2\%) & $\bm{8.6} \%( 9.1\%)$    \\
 Hamsterster &50.2\%(50.1\%)& 31.7\%(30.4\%) & 48.3\%(48.3\%) & 35.1\% (34.7\%) &50\% (50.1\%) &43.3\% (44.4\%)&  $\bm{34.8\%} (34.4\%)$  \\
 50 Women & $48.5\% (48.1\%)$  & $31.8\% (30.5\%)$ & $36.6\% (34.3\%)$ & $18.3\%(11.4\%)$ &52.5\%(52.7\%)& $16\% (18.6\%)$ & $\bm{12.8\%} (9.7\%)$      \\
 %Hateful Users & 31.4\%(31.1\%) & 46.3\% (48.6\%) & 39.6\% ( 41.5\%)& 47.8\%(47.3\%)&\% (\%) &$\bm{}\%(\%)$       \\
\bottomrule
\end{tabular}
\label{OurMethodVSBaseline}
\end{table*}

\textbf{Interference evaluation for contagion}:
%We compare $CMatch$ with the state-of-art methods to show the benefit of $CMatch$ in reducing causal effect estimation error in the presence of contagion.
We choose two $CMatch$ variants with low estimation errors: % in Fig. \ref{fig:cmatch_tradeoff} 
\textbf{TNM2 + MSS + TCM3} and \textbf{TNM2 + C + TCM2}, denoted by $CMatch_{{MCL}_{MSS}}$ and $CMatch_{{MCL}_{C}}$ respectively, 
and compare their causal effect estimation error with the baselines. The first method uses a simpler cluster weight assignment while the second one uses the expensive maximum weighted matching of nodes.
 Fig. \ref{fig:CausalPlot} shows that both variants of $CMatch_{MCL}$ get significantly lower error than other methods, especially in Citeseer and Cora with $75.5\%$ and $81.8\%$ estimated error reduction in comparison to $CBR_{reLDG}$ for \emph{e.p = edge-weight}. $CMatch_{{MCL}_{MSS}}$ has higher error than $CMatch_{{MCL}_{C}}$ in most of the experiments which is expected as shown in Figure \ref{fig:cmatch_tradeoff}.
 % because of applying maximum weighted matching in node matching step, it is more expensive than the other variant.
$Randomized$ and $Match$ approaches have similar performance in all datasets because of their similarity in node randomization approach.
We also notice that $CBR_{reLDG}$ has the highest estimation error in Hamsterster data 
%for both $e.p = 0.1$ and $e.p = 0.5$ 
which confirms that clustering has a significant effect on the unallowable spillover. Meanwhile, $CMatch_{reLDG}$ outperforms other baselines in some datasets (Citeseer) and but not in others (Hamsterster and 50 Women). In Citeseer, the $CR$ method %and in Hamsterster $CBR_{reLDG}$ 
gets the largest estimation error.
%It is not enough to say $CMatch$ is the best and there needs to be an in-depth analysis of what these results are showing.

%We also explore the impact of spillover probability on the causal effect estimation. 
Fig. \ref{fig:CausalPlot} also shows that the higher the unallowable spillover probability, the larger the estimation error but also the better our method becomes relative to the baselines. For example, by increasing the unallowable spillover probability from $0.1$ to $0.5$ in Citeseer, the estimation error increases from $0.005$ to $0.02$ for $CMatch_{{MCL}_{C}}$ and from $0.023$ to $0.086$ for $CBR_{reLDG}$. 

\textbf{Interference evaluation for direct interference}:
Fig. \ref{fig:directinterference} shows the difference between RMSE of different estimators over the presence of direct interference for \emph{e.p = edge-weight}. In four datasets, both variants of $CMatch_{MCL}$ get the lowest estimation error in comparison to baseline methods. For example, $CMatch_{{MCL}_{C}}$'s error is approximately half of the error of $CBR_{reLDG}$ ($0.06$ vs. $0.13$ for Citeseer, $0.1$ vs. $0.22$ for Cora, $0.31$ vs. $0.54$ for Hamsterster, $0.15$ vs. $0.36$ for 50 Women). 
%After $CMatch_{MCL}$, $CMatch_{reLDG}$ gets the least error in comparison to other methods in all datasets except Citeseer. 
Similar to contagion, $Match$ and $Randomized$ methods have similar estimation errors.

\textbf{Potential spillover evaluation}:
Table \ref{OurMethodVSBaseline} shows the potential spillover %based on the number of edges and the sum of edge weights 
between treatment and control nodes assigned by different methods. This applies to both contagion and direct interference.
$CMatch$ has the lowest sum of edges and edge weights between treatment and control nodes across all datasets. The difference between $CMatch_{{MCL}_{C}}$ and the baselines in Cora and Citeseer is substantial: $CMatch_{{MCL}_{C}}$ has between $13.5\%$ and $34.8\%$ lower number of edges between treatment and control across datasets.

\begin{figure}
\centering
  \includegraphics[width=1\linewidth]{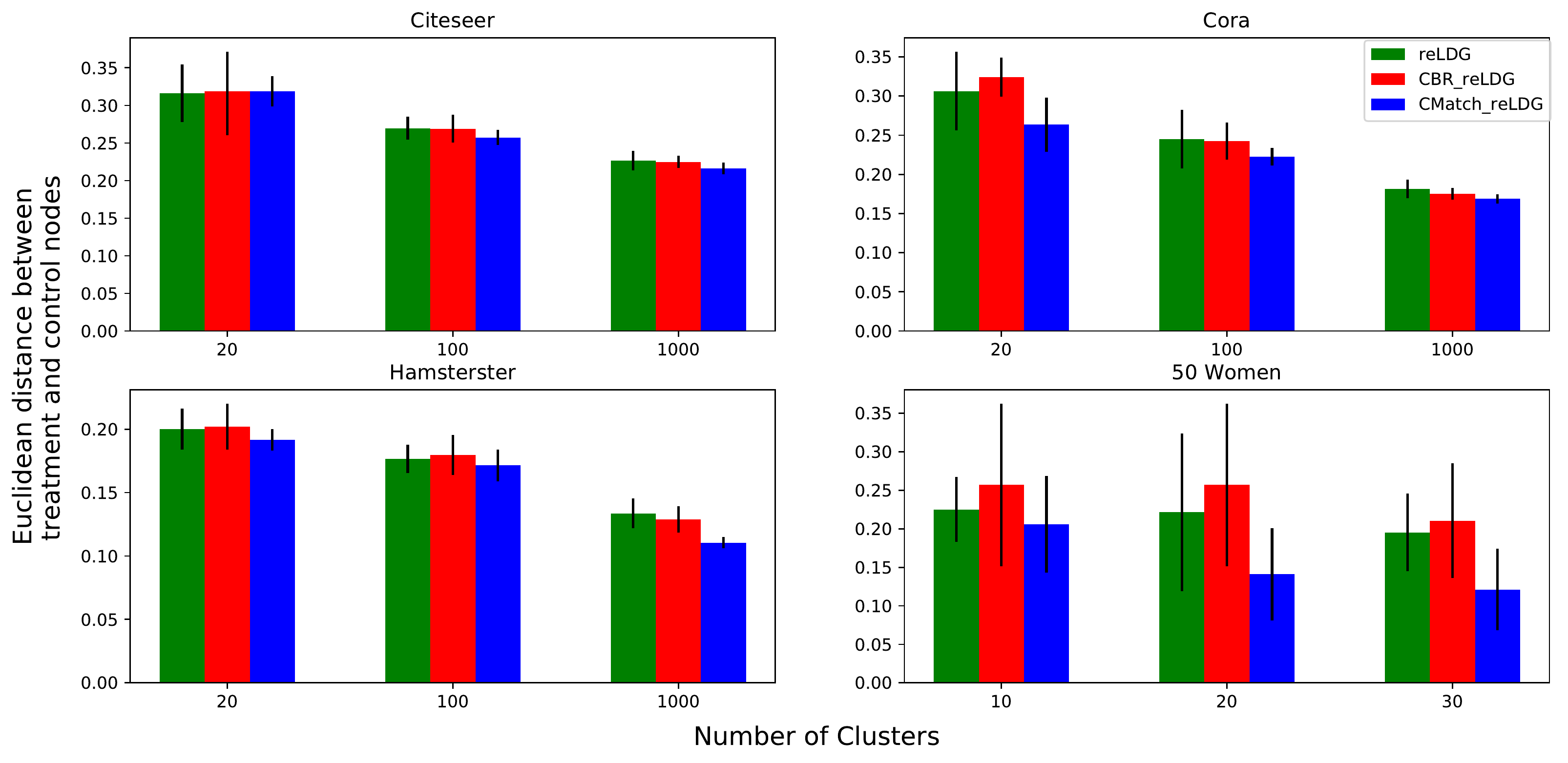}
\caption{Euclidean distance between the attribute vector means of treatment and control nodes for different number of clusters. The higher the number of clusters, the lower the selection bias.}
\label{fig:Attribute_distance}
%\vspace{-18pt}
\end{figure}

\textbf{Selection bias evaluation for contagion}. In this experiment, we look at the relationship between number of clusters and the difference between treatment and control nodes with and without cluster matching. Fig. \ref{fig:Attribute_distance} shows the Euclidean distance between the average of treatment and control nodes' attributes in $CMatch_{reLDG}$, $CBR_{reLDG}$ and $reLDG$ for three different number of clusters and unallowable spillover probability \emph{e.p = edge-weight}. 
Since $CMatch_{reLDG}$ optimizes for selection bias directly, it is not surprising that it results in treatment and control nodes that have more similar feature distributions than the other two methods. In Citeseer the differences are more subtle than in the other datasets.  
Error bars show the variance of averages over $10$ runs which confirms the low variance of estimations in all datasets except in 50 Women, which is a small dataset.

\begin{figure}
\centering
  \includegraphics[width=0.9\linewidth]{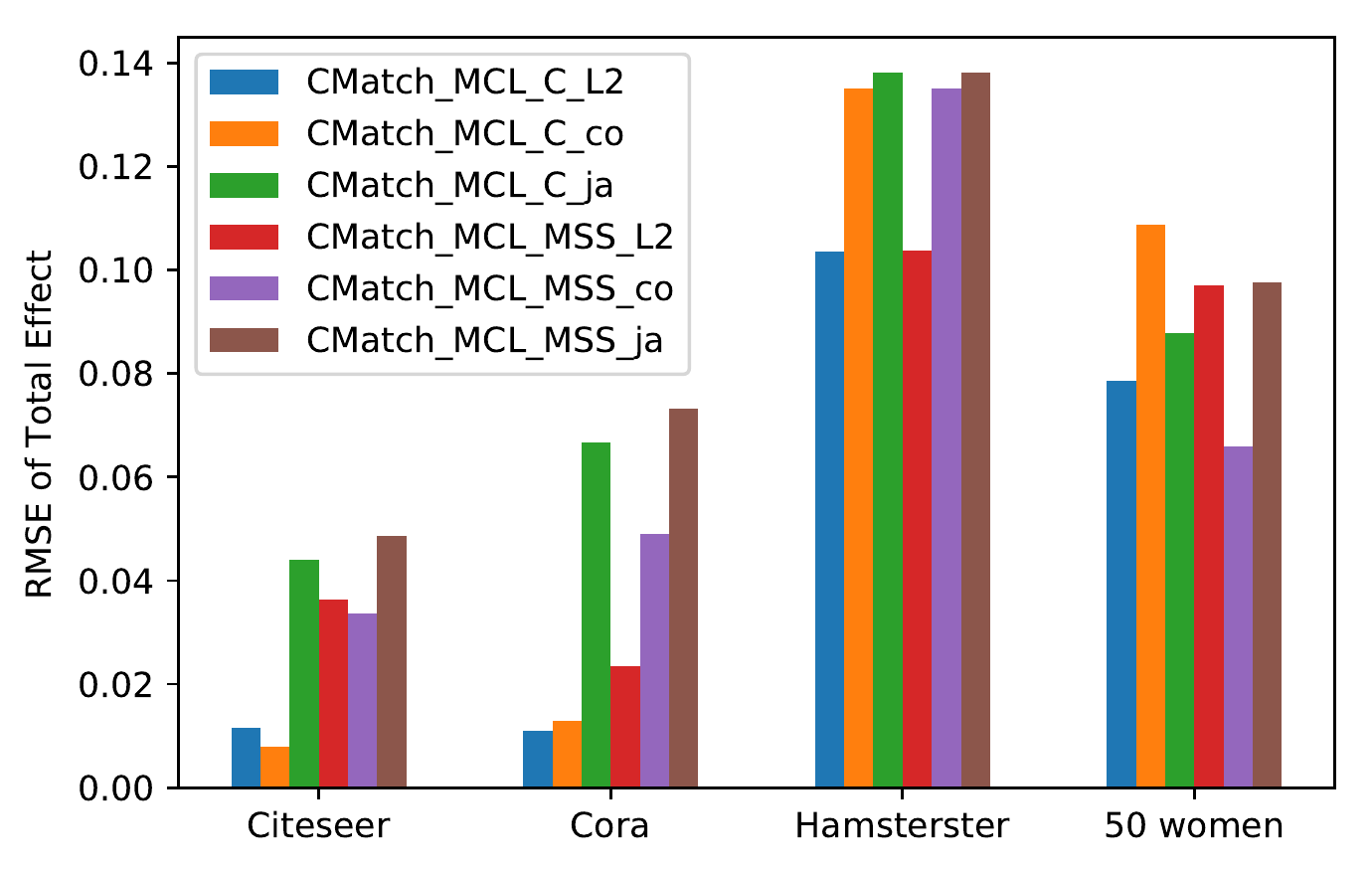}
\caption{RMSE of total effect in the presence of contagion using three different similarity methods to calculate spillover probability: Cosine (co), Jaccard (ja) and L2 similarity.}
\label{fig:similarity_metric}
\end{figure}

\textbf{Sensitivity to spillover probability metrics}. Our last experiment compares metrics for calculating the spillover probability, Cosine similarity, Jaccard similarity and the L2-based similarity used in all other experiments. We report on RMSE of total effect using $CMatch_{{MCL}_{C}}$ and $CMatch_{{MCL}_{MSS}}$ methods under contagion. Figure \ref{fig:similarity_metric} shows that $CMatch_{{MCL}_{C}}$ with L2-based similarity obtains the least error in all datasets except for Citeseer where Cosine similarity has slightly lower error. For $CMatch_{{MCL}_{MSS}}$, Cosine similarity has the lowest RMSE in Citeseer and 50 Women dataset, while Euclidean similarity has the lowest error in the other datasets. Jaccard similarity has the highest estimation error in all almost all cases. % except 50 Women.

\section{Conclusion}
\label{conclusions}

We presented $CMatch$, the first optimization framework that minimizes both interference and selection bias in cluster-based network experiment design. We demonstrated the tradeoff between causal effect estimation error and distance between treatment and control groups, as well as the value of combining weighted graph clustering with cluster matching. % and achieving much lower causal effect estimation error than state-of-the-art approaches. 
Our experiments on four real-world network datasets showed that $CMatch$
%considering weighted graph clustering can %reduce interference bias by $36\%$ to $74\%$ and 
reduces the causal effect estimation error by $8.6\%$ to $91.4\%$ when compared to state-of-the-art techniques. Some possible extensions of our framework include understanding the impact of network structural properties on estimation, jointly optimizing for interference and selection bias, and developing frameworks that are able to mitigate for multiple-hop diffusions.

\bibliographystyle{aaai}
\bibliography{references}
\end{document}